\documentclass[conference]{IEEEtran}
\IEEEoverridecommandlockouts

\usepackage{cite}
\usepackage{amsmath,amssymb,amsfonts}
\usepackage{algorithmic}
\usepackage{graphicx}
\usepackage{textcomp}
\usepackage{xcolor}

\usepackage{subcaption}
\usepackage{booktabs}
\usepackage{multirow}

\def\BibTeX{{\rm B\kern-.05em{\sc i\kern-.025em b}\kern-.08em
    T\kern-.1667em\lower.7ex\hbox{E}\kern-.125emX}}

\begin{document}
%
\title{Dynamic Prototype Rehearsal for Continual ECG Arrhythmia Detection}

\author{
\IEEEauthorblockN{Sana Rahmani\IEEEauthorrefmark{1}, Reetam Chatterjee\IEEEauthorrefmark{2}, Ali Etemad\IEEEauthorrefmark{1}, Javad Hashemi\IEEEauthorrefmark{2}
\IEEEauthorblockA{
\IEEEauthorrefmark{1}Department of Electrical and Computer Engineering\\
\IEEEauthorrefmark{2}School of Computing\\
Queen’s University, Kingston, Canada
\\\{sana.rahmani, 23vj22, ali.etemad, javad.hashemi\}@queensu.ca
}
}
}



\maketitle
\begin{abstract}

Continual Learning (CL) methods aim to learn from a sequence of tasks while avoiding the challenge of forgetting previous knowledge. We present DREAM-CL, a novel CL method for ECG arrhythmia detection that introduces dynamic prototype rehearsal memory. DREAM-CL selects representative prototypes by clustering data based on learning behavior during each training session. Within each cluster, we apply a smooth sorting operation that ranks samples by training difficulty, compressing extreme values and removing outliers. The more challenging samples are then chosen as prototypes for the rehearsal memory, ensuring effective knowledge retention across sessions. We evaluate our method on time-incremental, class-incremental, and lead-incremental scenarios using two widely used ECG arrhythmia datasets, Chapman and PTB-XL. The results demonstrate that DREAM-CL outperforms the state-of-the-art in CL for ECG arrhythmia detection. Detailed ablation and sensitivity studies are performed to validate the different design choices of our method.

\end{abstract}

\section{Introduction}
Electrocardiogram (ECG) signals are a crucial tool in clinical settings, widely used in applications such as arrhythmia detection \cite{Hannun2019}.
With the growing availability of new datasets from institutions worldwide, there is a pressing need for machine learning solutions that can adapt and learn from these evolving data sources, a process known as Continual Learning (CL). These solutions must not only acquire new knowledge (\textit{forward transfer}) but also retain previously learned information (\textit{backward transfer}), thus avoiding \textit{catastrophic forgetting}. In clinical settings, shifts in data distribution can exacerbate catastrophic forgetting, especially given the complex nature of ECG signals and cardiovascular diseases. Such shifts may arise from various factors, including patient demographics, disease types, recording equipment \cite{Kim2024, Ashhad2024}, and even the conditions under which the ECG is recorded, such as the time of day \cite{Kiyasseh2021}.

Various techniques have been proposed to address forgetting in the context of CL. Some methods approximate the distribution of data from previous sessions, either by storing a subset of past data in a memory buffer \cite{Kiyasseh2021} or by generating synthetic data with similar characteristics \cite{Cong2020}. Other approaches focus on the model itself, either through a deeper examination of the model parameters and their sensitivity to old and new data \cite{Chaudhry2018a} or analyzing the model predictions \cite{Rudner2022}. However, constraining a model to retain knowledge from past sessions can limit backward knowledge transfer \cite{Lin2022} while over-reliance on memory can lead to inefficient storage use. This highlights the importance of selecting the most effective subset of data for memory storage. The complexity of ECG signals and the diverse nature of cardiovascular diseases further complicate this challenge, underscoring the need for a CL approach that can dynamically adapt to new information while retaining essential knowledge from previous experiences.

In this paper, we propose \textbf{D}ynamic prototype \textbf{R}ehearsal for \textbf{E}CG \textbf{A}rrhyth\textbf{m}ia classification with \textbf{C}ontinual \textbf{L}earning (DREAM-CL), a novel CL method for ECG arrhythmia detection that leverages rehearsal based on a selective memory of representative samples to mitigate catastrophic forgetting. DREAM-CL focuses on dynamically selecting difficult samples based on the behavior of data during the training phase. 
To achieve this, we cluster the training data based on exhibited loss updates, which are then used to identify prototypes to represent the data in the memory. To compress the extreme values and reduce the impact of outliers, we apply the Lambert W transform, followed by selecting the difficult samples in the clusters based on training loss to populate the rehearsal memory. To thoroughly evaluate the performance of our method, we test it across three incremental scenarios: time-incremental, class-incremental, and for the first time in the context of ECG, lead-incremental. We use two widely used public datasets PTB-XL \cite{Wagner2020} and Chapman \cite{Zheng2020} and achieve state-of-the-art performances in comparison to prior works.

The main contributions of the paper can be summarized as follows.
(\textbf{1}) We introduce DREAM-CL, a novel CL method for ECG arrhythmia detection, which leverages a dynamic prototype rehearsal memory for the first time. 
(\textbf{2}) Specifically, DREAM-CL selects representative prototypes for the memory by clustering data according to their learning behavior during each training session. 
In each cluster, our method then performs a smooth sorting operation that ranks samples based on training difficulty, compressing extreme values and removing outliers. The harder samples are then selected as the prototypes for the rehearsal memory.
(\textbf{3}) We evaluate our method on various CL scenarios using two popular ECG arrhythmia datasets, demonstrating that our approach outperforms the state-of-the-art. 

\section{Related Work}
\label{sec:Related Work}

\noindent\textbf{Arrhythmia classification.}
Several studies have explored classification of ECG signals for arrhythmia detection using a range of machine learning algorithms. In \cite{Hannun2019}, a 34-layer CNN on was employed on single-lead ECG signals, achieving cardiologist-level arrhythmia classification performance. 
In \cite{Hou2019}, Long Short-Term Memory (LSTM) networks were integrated into autoencoders for ECG signal analysis, followed by a support vector machine for final classification. In \cite{Wang2019}, a multi-stage framework of convolutional and attention layers was proposed to detect arrhythmias by capturing intricate features across ECG signals. 
In \cite{Yao2020}, CNN, LSTM, and attention mechanisms were integrated into a single framework to extract both spatial and temporal information from multi-lead ECG signals for arrhythmia detection.  Later in \cite{Soltanieh2022,Soltanieh2023}, contrastive self-supervised learning was explored, offering promising generalization capabilities for in-distribution and out-of-distribution ECG representation learning.
Finally, CNNs and Transformers were combined in MCTnet \cite{Zhang2024} to capture both local patterns and long-range dependencies in ECG signals, leading to robust arrhythmia classification.  




\noindent \textbf{Continual learning.}
A number of different approaches have been proposed in the literature for effective CL. Rehearsal-based methods leverage a rehearsal memory to store a subset of past data or pseudo-samples, which is periodically replayed during training to mitigate catastrophic forgetting \cite{Cong2020, Gupta2020}. These methods typically focus on either random sampling or strategic selection of data samples for the memory \cite{Chaudhry2018a}. 
On the other hand, regularization-based approaches focus on the model itself and its parameters \cite{Chaudhry2018a, Rudner2022}, introducing constraints or penalties to prevent significant changes to important weights \cite{Gupta2020}. By selectively slowing down the update of key parameters, these methods aim to retain past knowledge while still allowing the model to learn from new data. 



CL on \textit{ECG signals} has been investigated in only a few prior works. The proposed method in \cite{Kiyasseh2021} utilized an effective memory buffer to perform CL across multiple incremental scenarios of time, institute, domain, and class. In \cite{Kim2024}, a CL strategy relied on well-known regularization-based CL techniques, namely LwF \cite{Li2017}, EWC \cite{Kirkpatrick2017}, and MAS \cite{Aljundi2018}. This work leveraged generative methods to estimate the distribution of previous training sessions, and explored CL for ECG in the context of multi-institute sessions.


\begin{figure}
    \centering
    \includegraphics[width=0.98\linewidth]{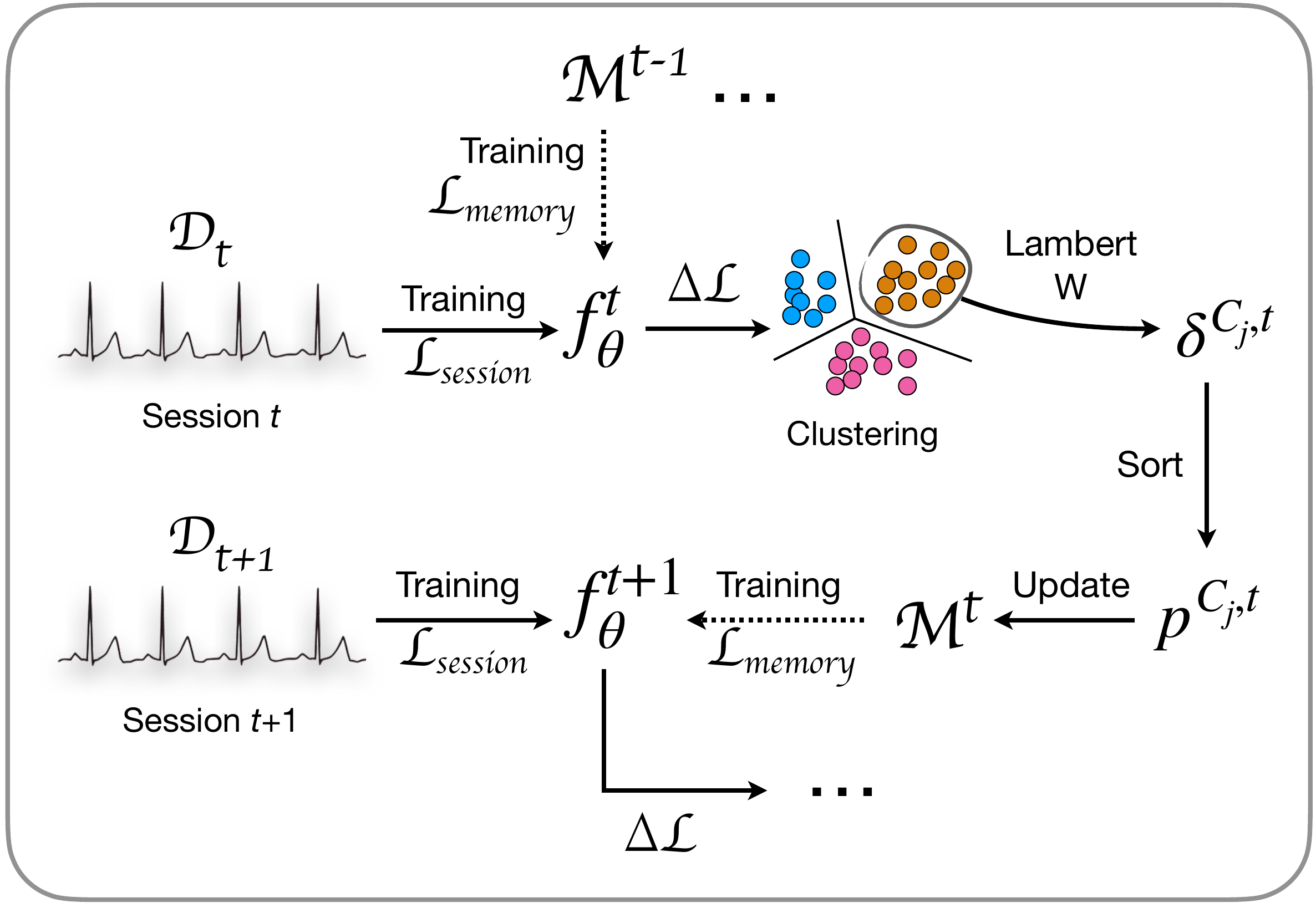}
    \caption{The overall framework of our proposed method.}
    \label{fig:framework}
\end{figure}

\section{Proposed Method}
\label{sec:methodology}

\noindent \textbf{Problem setup.} 
Let $\mathcal{D}_1, \mathcal{D}_2, \dots, \mathcal{D}_T$ represent a sequence of datasets, where each $\mathcal{D}_t = \{(x_i^t, y_i^t)\}_{i=1}^{N_t}$ consists of ECG signals $x_i^t \in \mathbb{R}^{d}$ and corresponding arrhythmia labels $y_i^t \in \mathcal{Y}$ used for training session $t$. Our goal is to train a model $f_\theta$, parameterized by $\theta$, to classify ECG samples into different arrhythmia types while continually training on the data from each $\mathcal{D}_t$. The objective is to minimize a loss function $\mathcal{L}(f_\theta(x), y)$ to not only improve performance on the current dataset $\mathcal{D}_t$ (forward transfer), but also to ensure that the performance on previous datasets, $\mathcal{D}_1, \dots, \mathcal{D}_{t-1}$, is not degraded (backward transfer). 







\noindent \textbf{Our approach.}
In our proposed method, DREAM-CL, we aim to build a rehearsal memory buffer by strategically selecting representative and challenging samples from each training session. Our approach captures the model's behavior during training by analyzing the changes in loss values over epochs, followed by clustering the samples based on their training dynamics, and selecting key samples to form the memory buffer for future learning sessions. An overview of our method is depicted in Fig. \ref{fig:framework}. Below we provide the detailed description of our method.

At each session $t$, we train $f_\theta$ using $\mathcal{D}_t = \{(x_i^t, y_i^t)\}_{i=1}^{N_t}$, and store the loss value for each $x_i^t$ over the training epochs. Next, we analyze the loss updates by calculating the difference in loss values between consecutive epochs by
\begin{equation}
\small
    \Delta \mathcal{L}(x_i^t, e) = \mathcal{L}_{e}(f_\theta(x_i^t), y_i^t) - \mathcal{L}_{e-1}(f_\theta(x_i^t), y_i^t)
\end{equation}
where $e$ denotes the epoch number and $\mathcal{L}_{e}$ is the corresponding loss value at epoch $e$. 
Accordingly, each data point $(x_i^t, y_i^t)$ can be represented by a vector $\nu_i^t = \{\Delta \mathcal{L}(x_i^t, e)\}^{m-1}_{e=0}$ where $m$ is the total number of training epochs. This vector effectively captures the impact of each sample on the model's training. Next, we group $\mathcal{D}_t$ based on $\{\nu_i^t\}_{i=0}^{N_t}$ to form clusters $C_1,C_2,\dots,C_k$, where each cluster contains samples with similar loss update patterns. Our objective is to populate the rehearsal memory buffer $\mathcal{M}_t$ with a representative distribution of prototype sets $p_1,\dots,p_k$ from corresponding clusters $C_1,C_2,\dots,C_k$, ensuring that different training behaviors are well-captured. 

Prior works have shown that harder samples are generally more informative for training \cite{Zhou2023}. Accordingly, we focus on the loss values of samples in each cluster to identify the prototypes. Additionally, samples with higher loss values are more likely to represent smaller classes which are more prone to catastrophic forgetting \cite{Bang2021}. However, focusing purely on the most difficult samples as per the loss values can be a risky strategy. While these samples may represent the most challenging cases for the model, they can also include outliers or noisy data points that could distort the learning process. Such outliers, if prioritized, may cause the model to overfit to incorrect or atypical examples, ultimately degrading generalization \cite{Kim2021}. To mitigate this, we apply the Lambert W function to the loss values before sorting to help compress extreme values, preventing the dominance of outliers and allowing for a more balanced selection of both hard and representative samples \cite{Castells2020}. This ensures that we obtain a smooth rehearsal memory buffer that captures diverse examples across varying levels of difficulty, thereby promoting a robust CL without overemphasizing noisy or anomalous data. Accordingly, we define
\begin{equation}
\small
d^{C_j, t}_i=\lambda \left( \mathcal{L}(x^{C_j, t}_i,y^{C_j, t}_i)-\mathcal{L}(\mathcal{D}^{C_j}_t) \right),
\end{equation}
where we normalize the final loss value for each sample in $C_j$ by subtracting it from the average loss of that cluster $\mathcal{L}(\mathcal{D}^{C_j}_t)$ and scaling the outcome by $0 < \lambda \leq 1$. The scaling has been shown to help the transform with better convergence. Finally, we measure the sample difficulty values based on
\begin{equation}
\small
\label{Eq:difficulty}
    \delta^{C_j, t}_i =e^{-W(0.5\times \max(-\frac{2}{e},d^{C_j, t}_i))}  
\end{equation}
where $W(x)$ is the solution to the equation $W(x)e^{W(x)} = x$. By sorting the values based on $\delta^{C_j, t}_i$, we select the top $|\mathcal{M}_t|/k$ examples from each cluster to form $\mathcal{M}_t$. Repeating this operation for each session, we replay the rehearsal memory $\{\mathcal{M}_j\}_{j=1}^{t-1}$ alongside $\mathcal{D}_t$ to train $f_\theta$
using the compound cross-entropy loss as follows
\begin{align}
\small
    \mathcal{L} = \dfrac{1}{N_t + |\{\mathcal{M}_j\}_{j=1}^{t-1}|} ( 
    \underbrace{\sum_{i=1}^{N_t} \sum_{z=1}^{c_t} y_{i,z}^t \log(f_\theta(x_{i,z}^t))}_{\mathcal{L}_\textit{session}} \nonumber \\ + \underbrace{\sum_{j=1}^{s-1} \sum_{i=1}^{\Tilde{|M_j|}} \sum_{z=1}^{c_t} y_{i,z}^j \log(f_\theta(x_{i,z}^j))}_{\mathcal{L}_\textit{memory}} 
    ),
\end{align}
where $c_t$ is the number of arrhythmia classes according to $\{\mathcal{D}_j\}_{j=1}^t$. By minimizing $\mathcal{L}_\textit{session}+\mathcal{L}_\textit{memory}$, the model is trained to improve its performance on the current dataset $\mathcal{D}_t$ while preserving knowledge from previous sessions.

\begin{table*}[t]
   \caption{The performance of our method across three scenarios of time, class and lead incremental compared to prior works.}
   \label{tab:performance}
   \large
\centering    
    \setlength{\tabcolsep}{10pt}
\resizebox{0.98\textwidth}{!}{   
   \begin{tabular}{l|ccc|ccc|ccc}
   \toprule
    \multirow{2}{*}{\textbf{Methods}} & \multicolumn{3}{c|}{\textbf{Time-incremental}}& \multicolumn{3}{c|}{\textbf{Class-incremental}}&\multicolumn{3}{c}{\textbf{Lead-incremental}}\vspace{1mm} \\
    & Mean $(\uparrow)$ & AvgAUC $(\uparrow)$ &  BWT $(\uparrow)$ &  Mean $(\uparrow)$ & AvgAUC $(\uparrow)$ &  BWT $(\uparrow)$&  Mean $(\uparrow)$ & AvgAUC $(\uparrow)$ &  BWT $(\uparrow)$\vspace{1mm}\\
   
   \hline
   
   CLOPS \cite{Kiyasseh2021} & 0.83{\scriptsize ±0.01} & 0.76{\scriptsize ±0.01} & \textbf{0.01{\scriptsize ±0.01}}
   
    & 0.70{\scriptsize ±0.03} & 0.71{\scriptsize ±0.02} & \underline{0.10{\scriptsize ±0.02}} 
    
    & 0.70{\scriptsize ±0.04} & \underline{0.70{\scriptsize ±0.04}} & 0.02{\scriptsize ±0.02}\vspace{1mm} \\
    
   A-GEM \cite{Chaudhry2018a} 
   & 0.80{\scriptsize ±0.00} & 0.72{\scriptsize ±0.01} & -0.06{\scriptsize ±0.02} 
   
   & 0.63{\scriptsize ±0.00} & 0.66{\scriptsize ±0.01} & 0.06{\scriptsize ±0.01}
   
   & 0.63{\scriptsize ±0.01} & 0.69{\scriptsize ±0.00} & \textbf{0.05{\scriptsize ±0.01}}\vspace{1mm} \\
   
   LAMAML \cite{Gupta2020} 
   & 0.82{\scriptsize ±0.01} & 0.68{\scriptsize ±0.03} & -0.05{\scriptsize ±0.02} 
   & 0.64{\scriptsize ±0.0} & 0.69{\scriptsize ±0.01} & 0.06{\scriptsize ±0.01}
   
   & 0.65{\scriptsize ±0.01} & \underline{0.70{\scriptsize ±0.00}} & \textbf{0.05{\scriptsize ±0.01}} \vspace{1mm}\\
   
   TFS \cite{Wu2023} & 0.78{\scriptsize ±0.01} & 0.64{\scriptsize ±0.01} & \underline{-0.04 {\scriptsize ±0.02}}
   
   & 0.62{\scriptsize ±0.01}& 0.62{\scriptsize ±0.01}& -0.01{\scriptsize ±0.01}
   
   & 0.65{\scriptsize ±0.02} & 0.65{\scriptsize ±0.02} & \textbf{0.05{\scriptsize ±0.01}}\vspace{1mm}\\
   \textbf{DREAM-CL ($r_\mathcal{M} = 0.25$)} &  \underline{0.85{\scriptsize ±0.01}} & \underline{0.78{\scriptsize ±0.01}} & \textbf{0.01{\scriptsize ±0.02}}
   
   & \underline{0.73{\scriptsize ±0.02}} & \underline{0.73{\scriptsize ±0.02}} & \underline{0.10{\scriptsize ±0.02}}
   
   & \underline{0.72{\scriptsize ±0.02}} & \textbf{0.72{\scriptsize ±0.02}} & \underline{0.03{\scriptsize ±0.01}}\vspace{1mm}\\

   \textbf{DREAM-CL ($r_\mathcal{M} = 0.75$)} &  \textbf{0.86{\scriptsize ±0.01}} & \textbf{0.79{\scriptsize ±0.01}} & \textbf{0.01{\scriptsize ±0.02}}
   
   & \textbf{0.74{\scriptsize ±0.01}} & \textbf{0.75{\scriptsize ±0.01}} & \textbf{0.11{\scriptsize ±0.02}}
   
   & \textbf{0.73{\scriptsize ±0.01}} & \textbf{0.72{\scriptsize ±0.02}} & \textbf{0.05{\scriptsize ±0.01}}\vspace{1mm}\\
   \bottomrule
   \end{tabular}
   }
   \label{tab:ablation}
\end{table*}

\begin{figure*}[t] 
    \centering
    \begin{minipage}{\textwidth} 
        \centering
        \begin{subfigure}{0.32\textwidth}
            \centering
            \includegraphics[width=\linewidth]{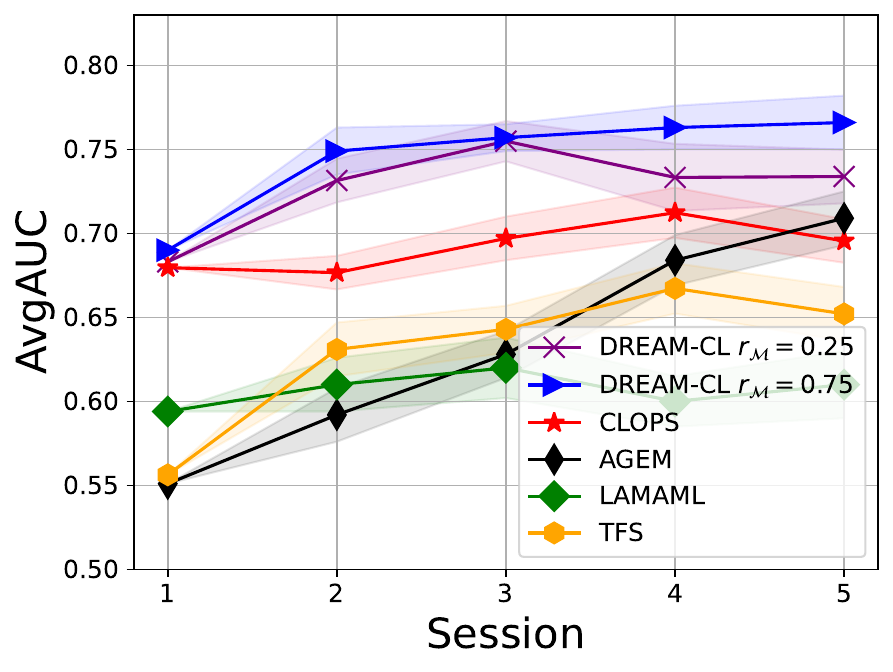}
            \caption{Class-incremental}
		\label{fig:sessprogclass}
        \end{subfigure}
        \hfill
        \begin{subfigure}{0.32\textwidth} 
            \centering
            \includegraphics[width=\linewidth]{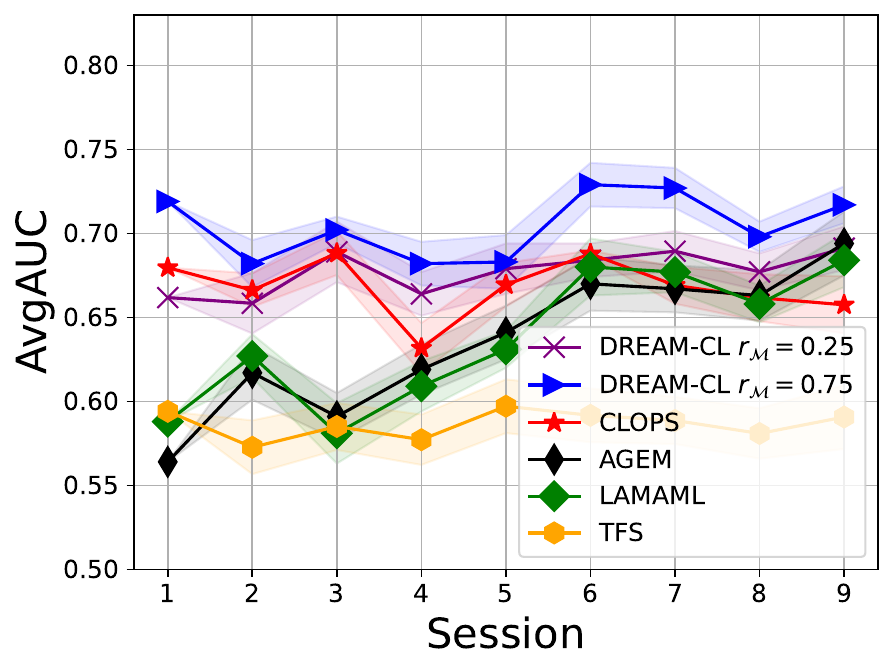} 
            \caption{Lead-incremental}
		\label{fig:sessproglead}
        \end{subfigure}
        \hfill
        \begin{subfigure}{0.32\textwidth} 
            \centering
            \includegraphics[width=\linewidth]{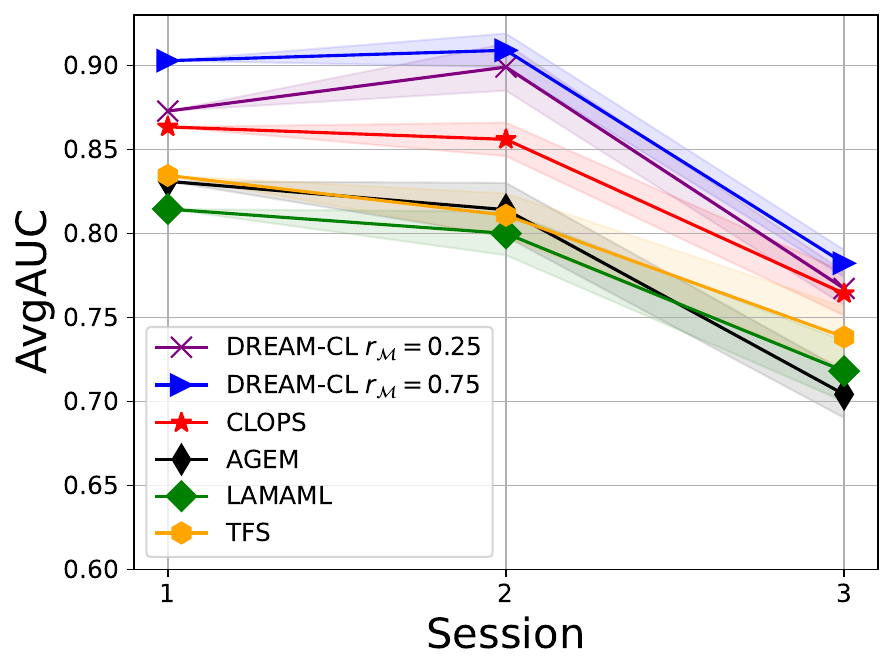} 
            \caption{Time-incremental}
		\label{fig:sessprogtime}
        \end{subfigure}
        \vspace{-2mm}
        \caption{Performance during training on different sessions.}
	\label{fig:SessionProg}
    \end{minipage}
\end{figure*}


\section{Experiments}

\noindent\textbf{Datasets.} We use the following datasets to evaluate our method on different CL scenarios: 
(\textbf{1}) \textbf{Chapman} \cite{Zheng2020} consists of 10,464 patient recordings of 12-lead ECG. Each data point is labeled with one out of 
four arrhythmia classes. (\textbf{2}) \textbf{PTB-XL} \cite{Wagner2020} includes 18,885 patient recordings of 12-lead ECG, across 5 arrhythmia classes.

\noindent\textbf{CL scenarios.}
We conduct experiments across three CL scenarios: time-incremental, class-incremental, and lead-incremental. Following \cite{Kiyasseh2021}, we consider the time tags reported in Chapman to define three sessions for time-incremental. As for class-incremental, we divide PTB-XL into five sessions based on the five classes present in the dataset, incrementally including more classes in each session. Finally, for lead-incremental, we consider PTB-XL and divide it into nine sessions, with each session including a random subset of leads.

\noindent\textbf{Baselines.}
We compare our method to CLOPS \cite{Kiyasseh2021}, A-GEM \cite{Chaudhry2018a}, LAMAML \cite{Gupta2020}, 
and TFS \cite{Wu2023}. Given the scarcity of prior works on the specific topic of CL for ECG, we adapted state-of-the-art CL methods A-GEM and LAMAML (originally proposed for CL in other domains), for the task of arrhythmia detection. We produced all the baseline results to comply with our experiment setup for a fair comparison.

\noindent\textbf{Evaluation criteria.} 
In the final session $t$, we report the average AUC score as $\text{AvgAUC} = \frac{1}{t} \sum_{j \leq t}{\text{AUC}}(f_{\theta}^t(\mathcal{D}_j^{\textit{test}})),$ where the AUC score is computed for the model's performance on the test data $\mathcal{D}_j^{\textit{test}}$ for each session $j$. The backward transfer (BWT) is defined as
\begin{equation}
\footnotesize
    \text{BWT}=\frac{1}{t-1} \sum_{j=1}^{t-1} \left(\text{AUC}(f_{\theta}^t(\mathcal{D}_j^{\textit{test}})) - \text{AUC}(f_{\theta}^j(\mathcal{D}_j^{\textit{test}})) \right),
\end{equation}
following \cite{Deng2021}. Negative values of BWT indicate forgetting of knowledge learned in previous sessions. Furthermore, we report the mean of the AvgAUC as
\begin{equation}
    \footnotesize
    \text{Mean}=\frac{1}{t} \sum_{j \leq k} \frac{1}{j} \sum_{t \leq j}\text{AUC}(f_{\theta}^j(\mathcal{D}_t^{\textit{test}})),
\end{equation} 
as proposed by \cite{Rebuffi2017}. This metric provides insight into the consistency of the model's performance across sessions with learning new tasks without affecting prior performance. All metrics are averaged across \textbf{five runs} with random seeds.

\noindent \textbf{Implementation details.} For the backbone of our model, we use a simple CNN commonly used by prior works on ECG-based CL \cite{Kiyasseh2021}. The architecture includes three blocks where each block contains one convolution layer followed by batch normalization, ReLU activation, maxpooling, and dropout. We perform our experiments on an NVIDIA GeForce RTX 2080 Ti GPU, implement our models with Pytorch, and use Adam for optimization.

\begin{table}
    \caption{Ablation study.}
    \centering
     \setlength{\tabcolsep}{4pt}

   \large
    \resizebox{0.9\columnwidth}{!}{   
    \begin{tabular}{lccc}
    \toprule
       Methods  & Mean & AvgAUC & BWT\\
       \midrule
       DREAM-CL  & {0.73{\scriptsize ±0.02}} & {0.73{\scriptsize ±0.02}} & {0.10{\scriptsize ±0.02}}\vspace{1mm}\\
       w/o kmeans (w/ GMM)  & 0.72{\scriptsize ±0.01} & 0.71{\scriptsize ±0.01} & 0.07{\scriptsize ±0.01}\vspace{1mm}\\       
       w/o Lambert W  & 0.72{\scriptsize ±0.02} & 0.73{\scriptsize ±0.01} & 0.05{\scriptsize ±0.01}\vspace{1mm}\\
       w/o $\delta$ (w/ random samples)  & 0.70{\scriptsize ±0.02} & 0.70{\scriptsize ±0.03} & 0.01{\scriptsize ±0.02}\vspace{1mm}\\
       w/o clustering (w/ random samples)& 0.68{\scriptsize ±0.03}&0.69{\scriptsize ±0.02}& 0.05{\scriptsize ±0.02}\vspace{1mm}\\
       \bottomrule
    \end{tabular}
    }
    \label{tab:my_label}
\end{table}

\noindent \textbf{Results.} 
In Table \ref{tab:performance} we present the performance of our model in comparison to state-of-the-art methods on three CL setups, namely time-incremental, class-incremental, and lead-incremental learning. We utilize two different rehearsal memory sizes for our method, small ($r_{\mathcal{M}} = 0.25$) and large ($r_{\mathcal{M}} = 0.75$), where $r_{\mathcal{M}} = |\mathcal{M}|/|\mathcal{D}$|. We observe that our model outperforms prior works, setting a new state-of-the-art. Among the competing methods, CLOPS demonstrates the overall second-best performance. Next, for a more detailed view of the performance of DREAM-CL, we report the AvgAUC across each session in Fig. \ref{fig:SessionProg}. Here, we observe that our method consistently exhibits strong performances throughout the incremental sessions. By studying the type of incremental sessions used to evaluate our method, we observe that lead-incremental seems more challenging for CL as evidenced by the fluctuations, which could be due to the challenging nature of adapting to different random sets of ECG leads.

Next, we perform a detailed ablation study, where we drop key components of our method and replace them with alternatives in the class-incremental scenario. First, we explore the impact of the clustering technique in our approach. As expected, the choice of segmentation strategy does not have a significant impact on the final performance, although kmeans does show slightly better results. Next, instead of applying the Lambert W transform and selecting the most difficult samples, we simply select the samples with the highest $\Delta \mathcal{L}$. Here, we observe that accounting for outliers or
noisy data through compressing extreme values using this technique improves the quality of the rehearsal memory, especially for backward transfer. We then explore two additional sample selection strategies where we perform \textit{random} sampling on two levels, with and without clustering. In this experiment, we observe that random sampling either with or without clustering yields lower performances in comparison to DREAM-CL.

\begin{table}
    \centering
        \caption{The effect of rehearsal memory size.}
        \large
\resizebox{0.9\columnwidth}{!}{       
    \begin{tabular}{cccccc}
    \hline
         $r_{\mathcal{M}}$ & 0.25 & 0.35 & 0.5 & 0.65 & 0.75\vspace{1mm}\\
    \hline\hline
         Class IL& 0.73{\scriptsize ±0.01} & 0.72{\scriptsize ±0.02} & 0.73{\scriptsize ±0.01} & 0.73{\scriptsize ±0.02} & 0.74{\scriptsize ±0.01}\vspace{1mm}\\
         Time IL& 0.85{\scriptsize ±0.01} & 0.86{\scriptsize ±0.0} & 0.86{\scriptsize ±0.01} & 0.85{\scriptsize ±0.0} & 0.86{\scriptsize ±0.01}\vspace{1mm}\\
         \hline
    \end{tabular}
    }
    \label{tab:memory}
\end{table}

\begin{figure}
    \centering
        \includegraphics[width=0.7\linewidth]{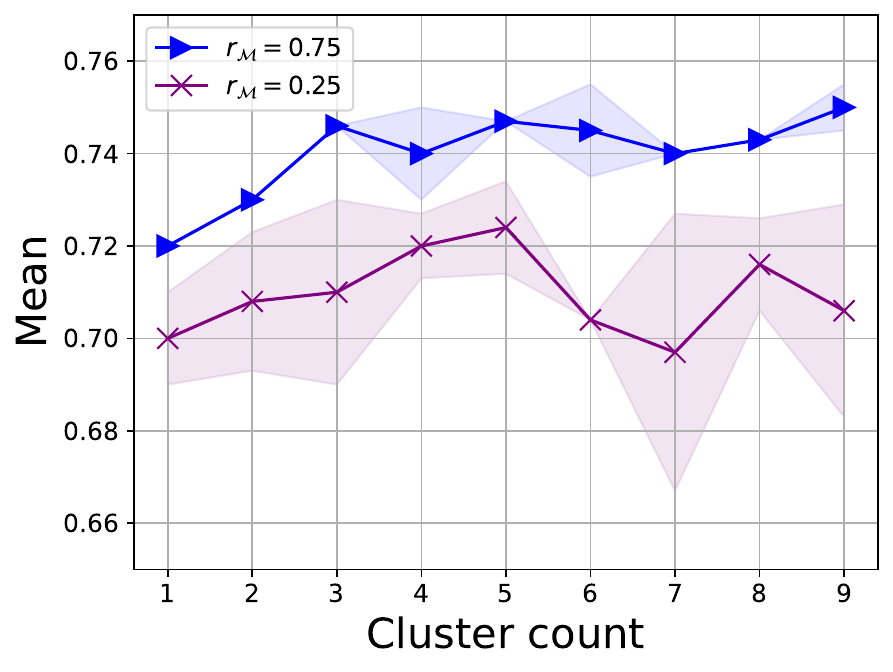} 
        \vspace{-2mm}
        \caption{Impact of no. clusters.}
        \label{fig:kmeans}
\end{figure}

In Table \ref{tab:memory}, we explore the impact of the size of the rehearsal memory buffer on our method, where we observe that the memory size does not have a strong impact on the performance of our method, indicating the effectiveness of our sample selection strategy. Finally, we perform a sensitivity study on the number of clusters used to identify the key samples for the rehearsal memory and present the results (class-incremental scenario) in Fig. \ref{fig:kmeans}. In this figure we do not observe a dominant trend beyond a the first few clusters. In our experiments, we set $k=5$ for all setups.

\section{Conclusion}
In this paper, we propose a new CL method for ECG arrhythmia detection. Our method selects representative prototypes for a rehearsal memory by clustering data based on their learning behavior and ranking samples by training difficulty, prioritizing harder examples. Extensive experiments validate the effectiveness of our method against the state-of-the-art.

\vspace{1mm}
\section*{Acknowledgment} We thank NSERC, Mitacs, and Ingenuity Labs Research Institute for partially funding this research.

\break

\bibliography{IEEEbib}
\bibliographystyle{IEEEbib}

\end{document}